\documentclass[letter]{ieice}
\usepackage[dvipdfmx]{graphicx,xcolor}
\usepackage[fleqn]{amsmath}
\usepackage{newtxtext}
\usepackage[varg]{newtxmath}

\usepackage{cite}
\usepackage{enumerate}
\usepackage{url}
\usepackage{subfigure}
\usepackage{here}
\usepackage{multirow}
\usepackage{CJKutf8} 

\usepackage{bm}
\usepackage{amsmath}
\usepackage{comment}

\usepackage{bbm}

\usepackage{algorithm}
\usepackage{algpseudocode}

\newcommand{\AmSLaTeX}{%
$\mathcal A$\lower.4ex\hbox{$\!\mathcal M\!$}$\mathcal S$-\LaTeX}
\def\BibTeX{{\rmfamily B\kern-.05em
\textsc{i\kern-.025em b}\kern-.08em
T\kern-.1667em\lower.7ex\hbox{E}\kern-.125emX}}
\makeatletter
\def\tmpcite#1{\@ifundefined{b@#1}{\textbf{?}}{\csname b@#1\endcsname}}%
\makeatother

\field{D}
\title{%
Self-supervised Neural Architecture Search for Multimodal Deep Neural Networks
}
\authorlist{%
 \authorentry{Shota Suzuki}{n}{labelA}\MembershipNumber{}
 \authorentry{Satoshi Ono}{m}{labelA}\MembershipNumber{0531898}
}
\affiliate[labelA]{The author is with the 
\EICdepartment{\texttt{%
Graduate School of Science and Engineering}}
\EICorganization{\texttt{Kagoshima University}}
\EICaddress{\texttt{Korimoto 1-21-40, Kagoshima, 890-0065 Japan}}
}

\received{2015}{1}{1}
\revised{2015}{1}{1}

\def\Vec#1{\mbox{\boldmath $#1$}}

\newcommand{\jtextd}[1]{}

\begin{document}
\maketitle
\begin{summary}
Neural architecture search (NAS), which automates the architectural
design process of deep neural networks (DNN), has attracted increasing
attention.
Multimodal DNNs that necessitate feature fusion
from multiple modalities
benefit from NAS due to their
structural complexity;
however, constructing an architecture for multimodal DNNs through NAS
requires a substantial amount of labeled training data. 
Thus, this paper proposes a self-supervised learning (SSL) method for
architecture search of multimodal DNNs.
The proposed method applies SSL comprehensively for both the
architecture search and model pretraining processes.
Experimental results demonstrated that the proposed method
successfully designed architectures for DNNs from unlabeled training
data.
\end{summary}
\begin{keywords}
self-supervised neural architecture search, multimodal neural network, contrastive learning, differential architecture search
\end{keywords}

\section{Introduction}\label{cha:introduction}

With the ongoing advancements in computational performance,
a significant progression has been realized toward the enlargement
of 
deep neural networks (DNNs), and it
is
increasingly common to
utilize neural architecture search (NAS)
to design
their network
structures, as evidenced by the design approach
utilized for 
EfficientNet~\cite{tan2019efficientnet}.
Recently, the practical applications of multimodal DNNs, which combine distinct
modalities, e.g., text, image, and sound, have progressed.
This integration of various modalities
promises enhanced
performance; however, it also necessitates the design of network structures
tailored to the specific features of the target data and tasks
because a
suitable network configuration differs
among distinct tasks.
Thus, there is an increasing expectation for the potential of NAS to
identify optimal architectures in multimodal DNNs, with research
efforts being pursued
vigorously~\cite{Perez2019,funokiDMNASDifferentiableMultimodal2021}.

Concurrently, there
is increasing 
interest in self-supervised Larning (SSL)
for representation learning
In particular,
contrastive learning (CL),
which leverages the
similarities between samples to facilitate learning in latent spaces,
has been
employed
in numerous tasks
due to
its methodological simplicity
and impressive downstream task performance (beginning
with image
classification~\cite{chen2020simple,sun2023contrastive}).
Despite ongoing research
on
the integration of SSL with NAS,
to the best of our knowledge, few studies
have attempted 
to apply SSL to NAS in the
multimodal DNN context.

Thus,
this paper proposes a NAS method for multimodal DNNs, which
employs an SSL approach
to explore the
fusion model structure and  the
critical points of modality integration.
The proposed method builds upon the bilevel multimodal NAS
(BM-NAS)~\cite{yinBMNASBilevelMultimodal2022}, which  relies on
a gradient-based approach \cite{Liu2018}, and it incorporates SimCLR to
perform the architecture search without labeled training data.
The proposed method was evaluated experimentally on the
MM-IMDB dataset~\cite{arevalo2017gated},
and the results
demonstrated
that, using unlabeled training data, the proposed method can
construct network architectures that are comparable to those discovered by
conventional methods using labeled training data.

\begin{table}[t]
  \centering
  \caption{Relationship between the previous and proposed methods.}
  \label{tbl:cbm-nas_positioning}
  \begin{tabular}{l|ll}
    \hline
         & Unimodal & Multimodal \\
    \hline
    Supervised learning      & DARTS~\cite{Liu2018}   & BM-NAS~\cite{yinBMNASBilevelMultimodal2022} \\
    Self-supervised learning & SSNAS~\cite{kaplan2020self} & Proposed method \\ 
    \hline
  \end{tabular}
\end{table}

\begin{figure}[t]
	\centering
        \rotatebox{0}{\includegraphics[width=1.0\linewidth]{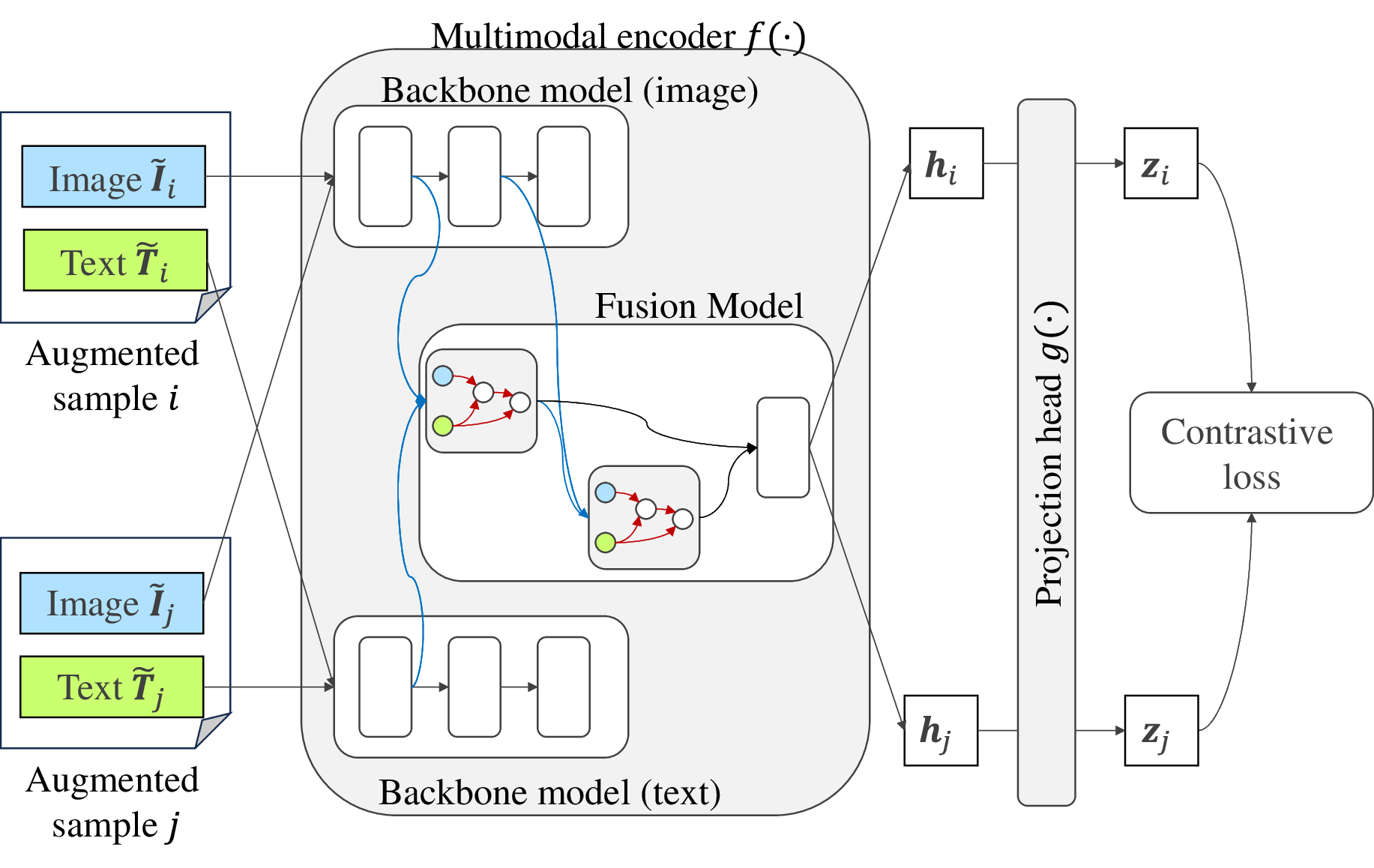}}
	\caption{Network structure and contrastive learning processes of the proposed method.}
	\label{fig:overview}
\end{figure}

\section{Related work}\label{cha:relatedWork}

Numerous studies have investigated
NAS, with more recent studies exploring
the introduction of gradient-based optimization strategies, e.g.,
DARTS~\cite{Liu2018}.
This has made it possible to
search for structures in an efficient manner 
by
converting the combinatorial search space into a continuous space,
thereby enabling gradient utilization.

DARTS has also been applied to the exploration of multimodal DNN
architectures~\cite{Perez2019,funokiDMNASDifferentiableMultimodal2021}.
For example, 
BM-NAS
proposed 
by Yin et al.~\cite{yinBMNASBilevelMultimodal2022} stands out for its
innovative use of directed pretrained networks for each modality,
exhibiting distinct advantages in terms of search cost and designed
model size.

As part of the ongoing innovations in the NAS field, the potential for SSL has also
been investigated~\cite{nguyen2021csnas, li2021bossnas}.
A notable advancement in this area is self-supervised neural
architecture search (SSNAS)~\cite{kaplan2020self}, which achieves NAS
through SSL by incorporating DARTS's search space into the SimCLR
framework.
SSNAS employs a methodology that, similar to SimCLR, computes loss by
addressing a pretext task of predicting augmented views, and then it
determines a network structure by optimizing weights using gradient
methods, thereby mirroring the strategy utilized in DARTS.

\section{Proposed method}\label{cha:proposed_method}

\subsection{Key concept}

The proposed method 
introduces SSL into the NAS of multimodal DNNs.
A comparison of 
the proposed method and
existing methods
is
given
in Table~\ref{tbl:cbm-nas_positioning}.
The proposed method builds on BM-NAS by adapting one-shot learning from
DARTS to multimodal DNNs, and it incorporates architecture search
leveraging SSL.
Specifically, the proposed method 
employs the contrastive loss introduced by SimCLR for
one-shot architecture search, thereby allowing for a search using only
unlabeled training data.

\subsection{Network architecture}

Fig.~\ref{fig:overview} shows that the proposed method,
which follows BM-NAS's paradigm, explores the architectures of DNNs for
multimodal data by merging backbone pretrained networks tailored to
each modality.
A multimodal encoder $f(\cdot)$ comprises backbone models and a
fusion model.
During contrastive learning, a projection head $g(\cdot)$ is attached
to the encoder, while a classification layer is used during supervised
learning.

Here, the
search space of the proposed method comprises the fusion model,
which is constructed from fusion cells that
function
to interconnect the
different modalities.
The architecture of the fusion model is determined by a
hierarchical framework
comprising
an upper level, which constructs a
directed acyclic graph made up of fusion cells and backbone networks,
and a lower level that identifies the types of inner step nodes within the
fusion cells and their linking ways.
The connections determined at the upper and lower levels are
shown as the 
blue and red lines in Fig.~\ref{fig:overview},
respectively.
The upper level addresses the issue of selecting features from the
backbone networks and determining the global structure of the fusion
model,
and the lower level focuses on learning the connections
between different modalities to create features
by determining 
specific algorithms.

When searching a network architecture,
fusion
cell candidates comprising
intermediate features from each modality's backbone
network are input, and
other fusion cells are output, both of which are weighted
by the parameter $\Vec{\alpha}$.
The internal structures of the fusion cells are determined by the inner step nodes
and their connections.
The input candidates for an inner step node include the outputs from two input
nodes and the other inner step nodes, which are combined by taking a
weighted sum with weight parameter $\Vec{\beta}$.
Here, each inner step node generates its output by applying a weight
paramter $\Vec{\gamma}$ to the outputs from five potential primitive operators.
Then, the output of the fusion cell is formed by concatenating
the outputs from all inner step nodes.

\subsection{Overview of architecture search}
\label{ssec:train}

Model training in the proposed method comprises three main steps: 
(1) an architecture search step utilizing CL;
(2) a representation learning step that continues to utilize CL after
establishing the architecture; and
(3) a classifier training step that uses a small amount of labeled data.

In step (1), guided by CL, the proposed method utilizes unlabeled training data to
explore the architecture.
This step is characterized by simultaneous learning of the network
architecture and its weights, with the final structure determined by
discarding the elements optimized to be of lesser significance.

After determining
the architecture, the removal of network
components leads to modifications in the outputs of the fusion cells and
the inner step nodes.
Thus, the proposed method mandates subsequent representation learning
and supervised learning, i.e., steps (2) and (3), representing post-architecture
finalization.
Note that the proposed method operates under the condition that labeled
data are exceedingly scarce; thus, it employs a
pretraining strategy that utilizes 
unlabeled training data
prior to
engaging in supervised learning with a minimal
amount
of labeled data for the downstream task.

\subsection{Architecture search using CL}
\label{sec:search}

Distinct from both DARTS and BM-NAS, the proposed method employs
CL
for the architecture search by optimizing
the weight parameters $\Vec{\alpha}$, $\Vec{\beta}$, and $\Vec{\gamma}$ attributed to
the components of the network.
During steps (1) and (2), the proposed method integrates a
projection head, i.e.,
a multilayer perceptron (MLP) layer
$g(\cdot)$, in the network
rather than the final classification layer.

Fig.~\ref{alg:ArchitectureSearchwithCL} shows the algorithm for
architecture search with CL.
The proposed method, which is  designed for multimodal DNNs, involves selecting
a
suitable data augmentation method tailored to each modality by feeding
the augmented data into the respective modality's backbone network.
First,
minibatch comprising $N$ samples is selected, and for
each instance
 $\Vec{x}_k = (\Vec{I}_k, \Vec{T}_k)$ 
in the minibatch, a random data
augmentation operation is applied to each modality to generate augmented view
sets 
$(\tilde{\Vec{I}}_i,\tilde{\Vec{T}}_i)$ and $(\tilde{\Vec{I}}_j,\tilde{\Vec{T}}_j)$.
For the image modality, an image 
$\Vec{I}_k$
 in $\Vec{x}_k$ undergoes operations selected
from available data augmentation techniques, e.g., cropping, flipping, color
adjustments, blurring, and rotating, to produce augmented image views 
$\tilde{\Vec{I}}_{k, i}$
 and 
$\tilde{\Vec{I}}_{k, j}$.
For the text modality, masking is applied to a text 
$\Vec{T}_k$ 
with a certain
probability to generate augmented text views 
$\tilde{\Vec{T}}_{k, i}$
 and
$\tilde{\Vec{T}}_{k, j}$.
Note that the data augmentation processes tailored to each modality
create both positive and negative instances, which are essential for
computing CL.

The augmented view sets are then applied to the network
during the architecture search to obtain representation vectors
$\Vec{h}_i$ and $\Vec{h}_j$.
These vectors are then input to the projection head $g(\cdot)$,
thereby mapping them to multimodal feature spaces as $\Vec{z}_i$ and
$\Vec{z}_j$.
Here,
CL $l(\Vec{z}_i, \Vec{z}_j)$ is calculated from
$\Vec{z}_i$ and
$\Vec{z}_j$, as follows:
\begin{flalign}
  l(\Vec{z}_i, \Vec{z}_j) = -\log \dfrac{{\rm exp}(sim(\Vec{z}_{i}, \Vec{z}_{j})/\tau)}
  {\sum\nolimits_{k=1}^{2N}
    \mathbbm{1}_{[k\neq i]}{\rm exp}(sim(\Vec{z}_{i}, \Vec{z}_{k})/\tau)}
\end{flalign}
where $\mathbbm{1}_{[k\neq i]}$, $sim(\cdot)$ and $\tau$ denote the
indicator function that returns $1$ if and only if $k \neq i$, cosine
similarity, and a temperature parameter, respectively.%
Previous studies
have demonstrated empirically 
that preparing a separate
space to apply CL (distinct from the encoder's feature space)
enhances the encoder's performance when utilized in downstream
tasks~\cite{chen2020simple}.

Based on the calculated loss, the proposed method alternates between
updating the architecture-determining weight parameters $\Vec{\alpha}$,
$\Vec{\beta}$, and $\Vec{\gamma}$, and the operator-specific weight parameter
$\Vec{w}$.
The updates for $\Vec{\alpha}$, $\Vec{\beta}$, and $\Vec{\gamma}$
leverage a validation dataset,
and the updates for $\Vec{w}$ utilize a
training dataset.

\begin{figure}[t]
	\centering
	\begin{algorithm}[H]
          \caption{Algorithm of architecture search for multimodal DNNs using contrastive loss}
		{\footnotesize	
		\begin{algorithmic}[1]
			\State Initialize $\Vec{\alpha}$, $\Vec{\beta}$, $\Vec{\gamma}$ and $\Vec{w}$
			\For {epoch $< T_e $}
			\For {phase $\in$ \{train, valid\}}
			\For {sampled minibatch $\{\Vec{x}_k\}^{N_{\rm phase}}_{k=1}$}
			\ForAll{$k\in \{1,...,N_{{\rm phase}}\}$} 
                        \State %
                               Generate augmented view sets 
                               $\tilde{\Vec{I}}_{{\rm phase}, 2k-1}$, $\tilde{\Vec{T}}_{{\rm phase}, 2k-1}$,
                               \hspace*{15mm} ~ 
                               $\tilde{\Vec{I}}_{{\rm phase}, 2k}$, and 
                               $\tilde{\Vec{T}}_{{\rm phase}, 2k}$ from
                               $\Vec{I}_{{\rm phase}, k}$ and $\Vec{T}_{{\rm phase}, k}$.
                        \State 
                               $\Vec{z}_{{\rm phase}, 2k-1} = g(f(\tilde{\Vec{I}}_{{\rm phase}, 2k-1}, \tilde{\Vec{T}}_{{\rm phase}, 2k-1} ))$
                        \State 
                               $\Vec{z}_{{\rm phase}, 2k} = g(f(\tilde{\Vec{I}}_{{\rm phase}, 2k}, \tilde{\Vec{T}}_{{\rm phase}, 2k} ))$
			\EndFor

                        \ForAll{$i \in \{1, \ldots, 2N_{\rm phase}\}$, $j \in \{1, \ldots, 2N_{\rm phase}\}$}
                          \State 
                                 $s_{i,j} = \Vec{z}_{{\rm phase},i}^{\rm T} \Vec{z}_{{\rm phase}, j} /
                                           (||\Vec{z}_{{\rm phase},i}|| ||\Vec{z}_{{\rm phase}, j}||   )  $
                        \EndFor
                        \State 
                               Define $l(i,j)$ as $l(i,j) = -\log 
                                \frac{\exp(s_{i,j}) / \tau}
                                {\sum_{k=1}^{2N_{\rm phase}} \mathbbm{1}_{[k\neq i]} \exp(s_{i, j}) / \tau }$
                        \State 
                               $ L = \frac{1}{2N} \sum_{k=1}^{N} [l(2k-1, 2k) + l(2k, 2k-1)] $
			\EndFor 
			\If {phase = train}
			\State Update $\Vec{w}$ 
                               based on 
                               $\nabla_{\Vec{w}}L$ 
			\Else
			\State Update $\Vec{\alpha},\Vec{\beta},\Vec{\gamma}$ 
                               based on 
                               $\nabla_{\Vec {\alpha},\Vec{\beta},\Vec{\gamma}}L$ 
			\EndIf
			\EndFor
			\If {higher validation accuracy is reached}
                        \State Store the best values of $\Vec{\alpha}, \Vec{\beta}, \Vec{\gamma}$
			\EndIf
			\EndFor
                        \State Derive the final architecture based on the best values of $\Vec{\alpha}, \Vec{\beta}, \Vec{\gamma}$
			\State \Return the derived architecture
		\end{algorithmic}
		}
	\end{algorithm}
        \vspace*{-5mm}
	\caption{Algorithm of the proposed architecture search using contrastive loss.}
	\label{alg:ArchitectureSearchwithCL}
\end{figure}

\begin{table}[t]
\centering
\caption{Number of samples used in Experiment 1.}
\label{tbl:training_data}
{\footnotesize
\begin{tabular}{l|ll}
\hline
              & Proposed method    & BM-NAS \\
\hline
Architecture & Unlabeled samples  & Labeled samples  \\
search        & $ 18,160 \times (1 - r)$ & $18,160 \times r$ \\
\hline
Representation & Unlabeled samples & ---                \\ 
learning       & $ 18,160 \times (1 - r)$ &         \\
\hline
Network weight & Labeled samples  & Labeled samples \\ 
training       & $18,160 \times r$ & $18,160 \times r$ \\
\hline
\end{tabular}
}
\end{table}

\begin{figure}[t]
	\centering
        \includegraphics[width=1.0\linewidth]{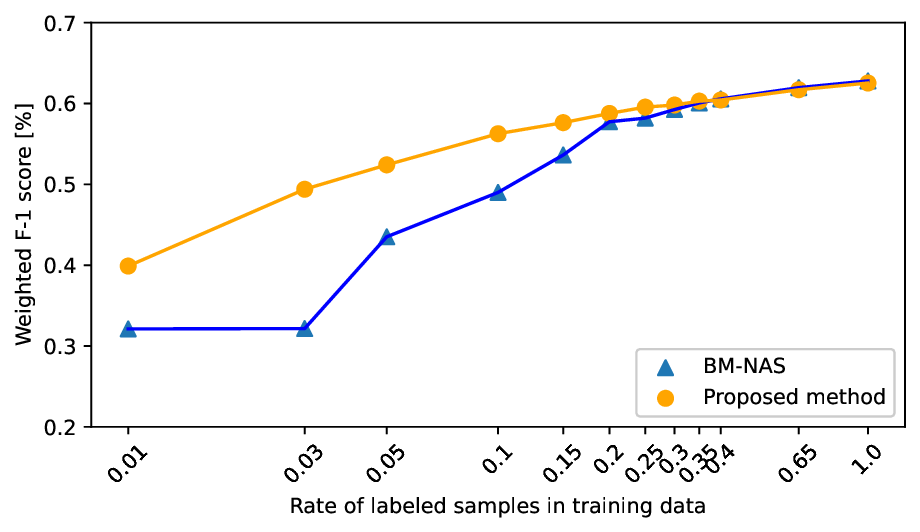}
	\caption{Comparison with the previous method.}
	\label{fig:result_reduce}
\end{figure}

\begin{figure*}[t]
	\centering
        {\scriptsize
        \begin{tabular}{@{}c@{~~~~~}c@{~~~~~}c@{}}
	\includegraphics[width=0.32\linewidth]{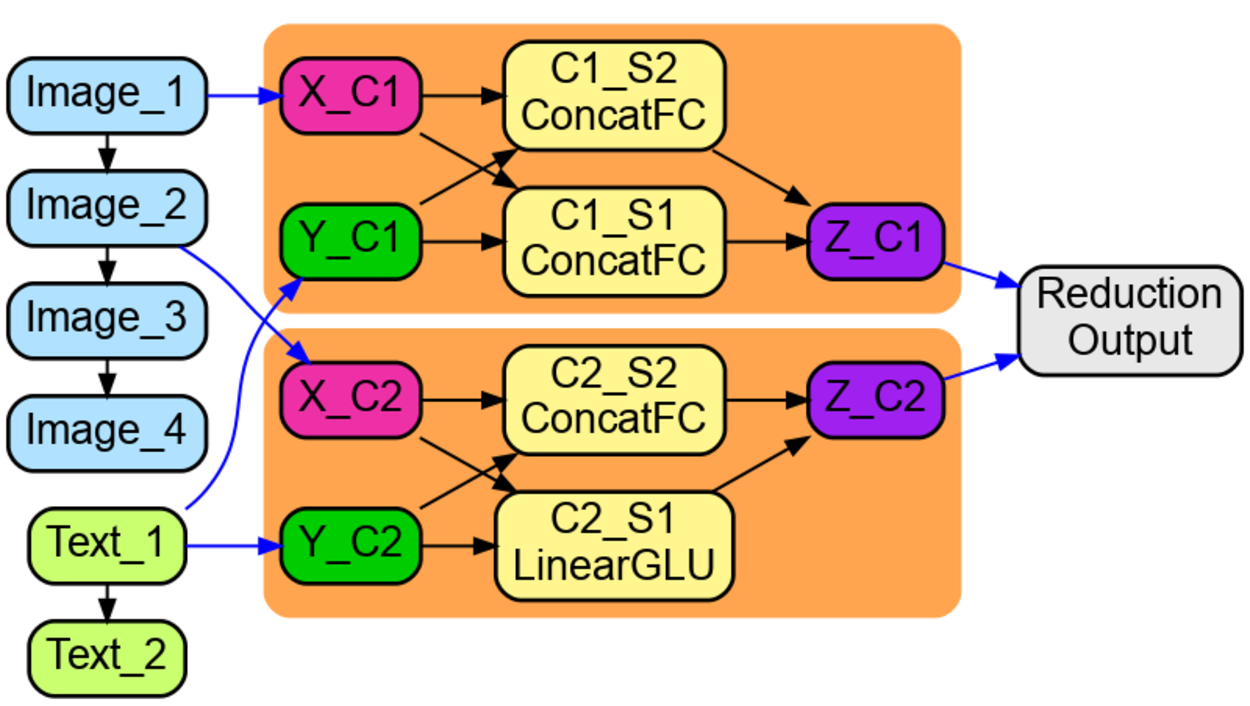} &
	\includegraphics[width=0.32\linewidth]{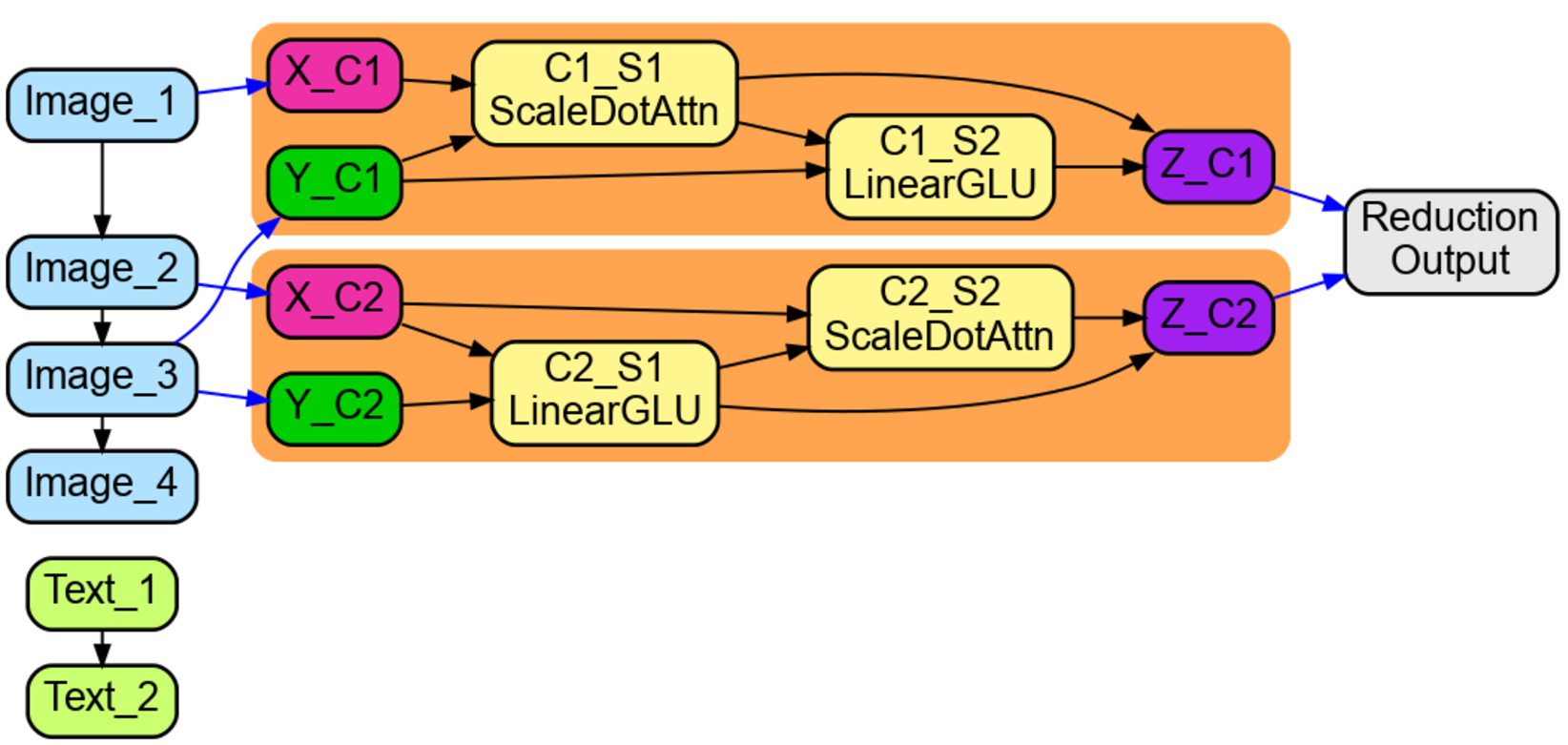} &

	\includegraphics[width=0.32\linewidth]{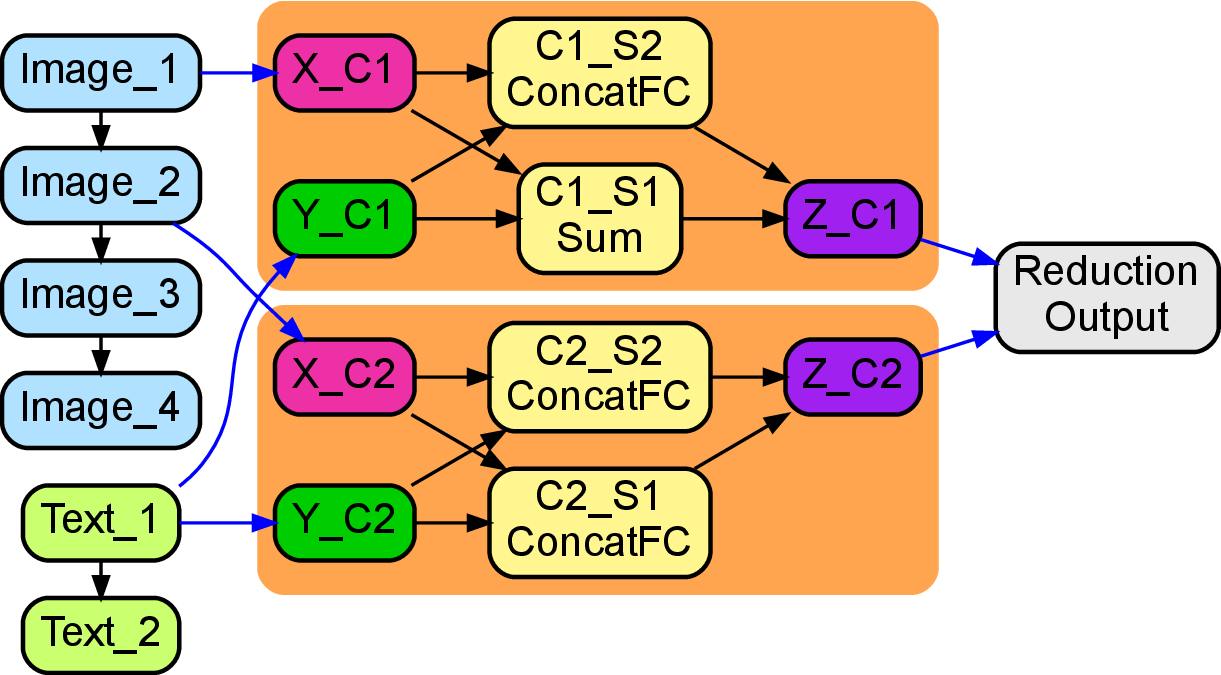} \\
        (a) Proposed method 
        &
        (b) BM-NAS ($r = 0.01$) &
        (c) BM-NAS ($r = 1.0$) 
        \end{tabular}
        }        
        \caption{Obtained network structures.}
        \label{fig:structure}
\end{figure*}

\subsection{Representation and classifier learning after architecture determination}

After finalizing the network architecture via the search method 
described in
Sec.~\ref{sec:search}, the model with the obtained structure undergoes
pretraining using the CL in the same manner employed in 
the architecture
search step.
Here, 
for each instance in the given batch, modality-specific random
augmentations are applied to form augmented view sets.
After these sets are input to the network, the weight $\Vec{w}$ are
updated based on the obtained CL loss.

Finally, to facilitate application of the pretrained model to the
downstream task, the classifier undergoes supervised training with a
limited set of labeled data.
In this step, the projection head $g(\cdot)$ used during the
CL steps is removed, and a classification layer is
added to the network.
The training data are input directly to the backbone network, and
multiclass cross-entropy loss is employed.

\section{Experimental evaluation}

The effectiveness of the proposed method was evaluated experimentally 
on the MM-IMDB dataset \cite{arevalo2017gated}, which was
designed for a multilabel classification task to determine movie
genres from images and textual information of movie posters.
The dataset comprises 23 classes and includes class imbalance.
In this evaluation, the maximum number of epochs $T_e$ was set to 30.
Here,
the weighted F1-score, which is an evaluation
metric that considers class imbalance, was considered in this experiment.

In Experiment 1, we compared the proposed method with
the existing
BM-NAS with
only a limited amount of labeled
training data.
Here, the proportion of labeled samples is denoted as $r$, and the value of $r$ varied from
0.01 to 1.0.
Table~\ref{tbl:training_data} shows the number of samples used for
training.
For consistency, we adopted the same quantity of training samples as in
the previous study~\cite{yinBMNASBilevelMultimodal2022}.
The proposed method used $18,160 \times (1 - r)$ unlabeled samples
during both the architecture search and representation learning processes, i.e.,
steps (1) and (2), and it used $18,160 \times r$ labeled samples
during the training of the downstream task, i.e., step (3).
In addition, 
BM-NAS used $18,160 \times r$ labeled samples for both the
architecture search and the model training process after the architecture
was determined.
Both the proposed method and BM-NAS employed the same pretrained
backbone models, i.e., Maxout MLP~\cite{goodfellow2013maxout} for text modality and VGG
Transfer~\cite{simonyan2014very} for image modality.
Despite their older architectures, these models were selected to
enable a straightforward and fair comparison with BM-NAS and other
multimodal DNNs on MM-IMDB
dataset~\cite{yinBMNASBilevelMultimodal2022}.

Fig.~\ref{fig:result_reduce} shows the averages of the weighted F1-scores
from 10 individual trials under each experimental condition.
As can be seen,
the score decreased for both methods
when the number of labeled
samples was reduced.
However, it is evident that the proposed method achieved higher scores
at lower levels of labeled samples $r \leq 0.3$, which confirms that the
proposed method's use of 
unlabeled training data for architecture search
and representation learning
is effective.

Figs.~\ref{fig:structure}(a) and (b)
show
examples of the architectures
obtained
using 
the proposed method and BM-NAS for $r = 0.01$,
respectively.
In addition,
Fig.~\ref{fig:structure}(c) shows an example discovered by BM-NAS
when $r = 1.0$.
The structures identified by the proposed method frequently included edges
leading to the fusion cell from both the first and second layers of
the image backbone network (i.e., Image\_1 and Image\_2), as well as the first layer
of the text backbone (i.e., Text\_1).
Furthermore, the inner step node frequently
selected
a concatenation layer
that simply connects the inputs as a single output.
Note that the structure shown in Fig.~\ref{fig:structure}(c) follows this trend,
which demonstrates that the proposed method
can generate
architectures that are similar to those produced by BM-NAS using a large amount 
of labeled data.
Conversely, when the $r$ vale was low, BM-NAS frequently derived inappropriate
structures that failed to utilize the text modality.

\begin{table}[t]
  \centering
  \caption{Experiment 2: Comparison on the designed network performance.}
  \label{tb:ex1}
  \begin{tabular}{@{}p{22mm}@{~}|@{~}c@{~~}l@{}}
    \hline
    Methods                                    & Approach & Weighted F1 [\%] \\
                                               &          & (standard error) \\
    \hline
    MFAS \cite{Perez2019}                      & Supervised      & $62.50$ \\
    BM-NAS \cite{yinBMNASBilevelMultimodal2022}& Supervised      & $62.92$ $\pm 0.03$ \\

    Proposed method                            & Self-Supervised & $62.46$ $\pm 0.06$ \\
    \hline
  \end{tabular}
\end{table}

In Experiment 2, to assess the performance of the architecture
discovered by the proposed method, we compared the discovered architecture with
those identified using the existing methods MFAS~\cite{Perez2019} and
BM-NAS~\cite{yinBMNASBilevelMultimodal2022}.
Here, the proposed method conducted the architecture search using only
unlabeled training data, without relying on any labeled data, while
the compared methods utilized labeled data.
After the architecture search, all the teste dmethods trained the obtained networks
using the labeled data.
In other words,
after architecture determination, the proposed method did not
engage in representation learning but optimized the weights $\Vec{w}$
through supervised learning using an equivalent amount of labeled
training data
used by the compared methods.

Table~\ref{tb:ex1} shows
the weighted F1-scores,
where the proposed method's result is averaged over 10 trials, while
the results of the other two methods are sourced from the previous
study~\cite{yinBMNASBilevelMultimodal2022}.
In this experiment, the proposed method primarily identified the
structure shown in Fig.~\ref{fig:structure}(a), and BM-NAS primarily
found the structure shown in Fig.~\ref{fig:structure}(c).
Table~\ref{tb:ex1} demonstrates that the proposed method, which employs
CL for its architecture search process, discovered
network architectures that are comparable to those found by MFAS and close to
the performance of those found by BM-NAS.

\section{Conclusion}

This paper proposed a gradient-based neural architecture search
method that employs self-supervised learning for multimodal neural networks.
The proposed method was evaluated experimentally on the
MM-IMDB dataset.
The results demonstrate that the
proposed method can discover model architectures that are comparable to those
found using existing supervised learning-based methods, even without labeled
data.
In the future, we plan to investigate additional experimental
validations on other datasets
to better understand the effectiveness of the proposed method.

\bibliographystyle{ieicetr}
\bibliography{reference}

\end{document}